\documentclass[11pt]{article}
\usepackage[T1]{fontenc}
\usepackage{lmodern}
\usepackage{cmap}
\usepackage[margin=1in]{geometry}
\usepackage{graphicx}
\usepackage{booktabs}
\usepackage{amsmath}
\usepackage{amssymb}
\usepackage{subcaption}
\usepackage{url}
\usepackage[numbers,sort&compress]{natbib}
\usepackage{hyperref}

\newcommand{\darkf}{\texttt{dark\_f}}

\title{Physical Kernel: Structured Visual Latents for Dark Manipulation}
\author{%
  Jinting Hang\textsuperscript{1,$\ast$},
  Hong Li\textsuperscript{1},
  Zhenhui Cai\textsuperscript{1},
  Zhihao Zhao\textsuperscript{1},
  and Jian He\textsuperscript{1}\\[0.6em]
  \small
  \textsuperscript{1}Harvest Praxis, China\\[0.45em]
  \textsuperscript{$\ast$}\texttt{jinting.hang@harvest-praxis.com}\\[0.15em]
  \texttt{hong.li@harvest-praxis.com}\\[0.15em]
  \texttt{zhenhui.cai@harvest-praxis.com}\\[0.15em]
  \texttt{zhihao.zhao@harvest-praxis.com}\\[0.15em]
  \texttt{jian.he@harvest-praxis.com}
}
\date{}

\begin{document}
\maketitle

\begin{abstract}
We study \emph{dark manipulation}: after a brief lit Write encodes $z_0=\mathrm{Enc}(\mathrm{rgb})$, a policy $\pi(z)$ and open-loop dynamics $f(z,a)$ complete contact-rich skills without further pixels (\darkf{}).
On ManiSkill StackCube ($n{=}160$; seed packs 0/1000), \darkf{} attains \textbf{68.1\% stacked} on the five-rung chain
(\texttt{near\_A}$\rightarrow$\texttt{grasped}$\rightarrow$\texttt{lifted}$\rightarrow$\texttt{on\_B}$\rightarrow$\texttt{stacked}),
compared with 35.6\% for per-step \texttt{lit\_reenc} and 0\% for \texttt{freeze}/\texttt{encode\_black}.
On a shared Write$\rightarrow$HOLD protocol ($n{=}40$), occlusion and camera-aligned GT contact-neighbor masks drive lit lift from 43\% to 0\%, while \darkf{} holds 82.5\%;
shuffling actions inflates dynamics MSE by ${\sim}9.4\times$;
write-time appearance shifts break encoding (night: 0\% stacked), yet the same shifts during HOLD leave \darkf{} lift unchanged;
Write length $T_w$ is flat once the stop phase is reached, while earlier stops and write-time blur/JPEG sharply cut stacked.
A dedicated $\pi_{\mathrm{write}}$ reaches 35\% vision-budget stacked ($n{=}80$); matched Dreamer-style/pixel nulls without privileged geom stay at 0\%.
Privileged state-RSSM MPC reaches ${\approx}35\%$ stacked with 9D dark observations---a stronger-observation null, not a matched visual baseline~\citep{hafner2020dreamer,ta2023maniskill3}.
\end{abstract}

\begin{figure}[t]
  \centering
  \includegraphics[width=\linewidth]{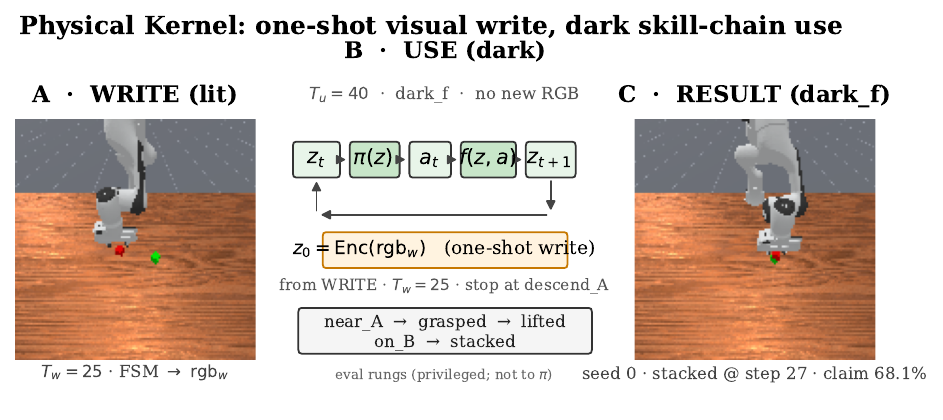}
  \caption{\textbf{Write--Use teaser (StackCube, seed~0).}
  \textbf{(A)~Write (lit):} privileged FSM ($T_w{=}25$) to \texttt{descend\_A}; one-shot $\mathrm{Enc}(\mathrm{rgb}_w)$.
  \textbf{(B)~Use (dark):} \darkf{} with $a_t=\pi(z_t)$, $z_{t+1}=f(z_t,a_t)$ for $T_u{=}40$ and no new RGB; rungs scored offline.
  \textbf{(C)~Result:} \texttt{stacked} on the same episode; aggregate claim 68.1\% ($n{=}160$).}
  \label{fig:protocol}
\end{figure}

\section{Introduction}

Robots that manipulate through contact must keep working when cameras drop frames, when hands or clutter occlude the scene, or when lighting and textures differ from training.
World models and JEPA-style predictors study representation quality under missing pixels~\citep{lecun2022path,assran2023self,bardes2024revisiting,hafner2020dreamer}, but they are often judged by generative fidelity or qualitative attention rather than by a fixed control protocol that mirrors deployment.

We focus on a concrete question: \emph{after one visual write}, can a structured physical latent finish a contact-rich skill chain \emph{in the dark}, with no further encoding?
If the answer is yes, open-loop latent dynamics and a latent policy can cover temporary sensing failures.
If not, continuous vision---or privileged state---remains necessary for late skills such as placing and stacking.

We build a \textbf{Physical Kernel}: an encoder $\mathrm{Enc}$, dynamics $f$, policy $\pi(z)$, and optional contact-aware updates, trained with lit distillation on ManiSkill StackCube~\citep{gu2023maniskill2,ta2023maniskill3,florence2020implicit}.\footnote{A companion paper~\citep{hang2026habit} asks \emph{why} freezing shared dynamics and adapting only a thin command map is preferable when demonstration multimodality reflects habit/action choice rather than multiple physical laws; the present manuscript focuses on Write--Use skill chains under \darkf{}.}
Evaluation separates \textbf{Write} (a privileged lit FSM to a fixed phase, then one encode) from \textbf{Use} (\darkf{}: $\pi$ and $f$ only).
We score \textbf{skill-chain rungs}, not video FID.
Sharing Write across conditions makes late-rung failures visible and keeps null comparisons fair.

\paragraph{Contributions.}
\begin{itemize}
  \item A locked protocol and deploy recipe (\texttt{litD4}+\darkf{}) with \textbf{68.1\% stacked} on the five-rung chain ($n{=}160$; packs 0 and 1000), beating per-step re-encoding and frozen latents on late rungs.
  \item Evidence that $f$ is action-sensitive (${\sim}9.4\times$ shuffle/true MSE) and that \darkf{} recovers HOLD lift under occlusion and GT contact masks where lit re-encode fails.
  \item An appearance-OOD split between \textbf{write-time} encoding fragility (dusk/night collapse) and \textbf{hold-time} robustness of \darkf{}, plus Write sensitivity: stacked is flat in $T_w$ once the stop phase is reachable, but drops sharply for earlier stops.
  \item Matched-Write nulls---LA stubs, freeze/black/scramble, and privileged state-RSSM MPC---with explicit rules so HOLD lift is not compared across unequal DARK information.
  \item Camera-projected GT contact masks (v2) with precision/recall and neighbor-vs-far checks on HOLD and skill-chain Use~\citep{bakhtin2019phyre}.
\end{itemize}

We do not claim stable ${\geq}75\%$ stacked on this recipe, SOTA over large video world models, or zero-shot transfer.
Hardware and second-domain Push results remain parallel tracks.

\section{Method}
\label{sec:method}

\subsection{Physical Kernel}
\label{sec:kernel}

The \textbf{Physical Kernel} is a visual encoder, a latent dynamics model, and a policy on a structured physical latent $z_p$ (denoted $z$ when clear).
From $128{\times}128$ RGB, $\mathrm{Enc}$ outputs $(z_p,z_a)$ with $\dim(z_p){=}64$ and $\dim(z_a){=}32$; control and open-loop roll use $z_p$ only.
Dynamics $f_\theta$ predict $\hat z_{t+1}=f(z_t,a_t)$.
The policy maps $z_t$ to $a_t=\pi(z_t)\in\mathbb{R}^4$ (3D end-effector delta and gripper).
Optional \textbf{ContactWrite} $CW(z,q)$ updates $z$ with proprioceptive features $q\in\mathbb{R}^5$.
Unless an ablation says otherwise, deployment uses \darkf{} without online $CW$.

\paragraph{Training.}
We train on 500 successful StackCube expert episodes (${\sim}18$k transitions) with discrete appearance domains (\texttt{default}/\texttt{warm}/\texttt{cool}/\texttt{dim}) and continuous color jitter.
The loss combines (i)~one-step dynamics consistency on $z_p$, (ii)~alignment of $z_p$ to a 9D privileged pose as a regularizer, (iii)~same-physics separation of $z_p$ across appearance pairs with domain prediction on $z_a$, and (iv)~behavioral cloning $\mathcal{L}_{\mathrm{BC}}=\|\pi(z_p)-a\|_2^2$.
Lit distillation~\citep{florence2020implicit} uses a privileged geometric FSM to provide target actions on rendered RGB; $\pi$ learns to match them from $z=\mathrm{Enc}(\mathrm{rgb})$ alone and never reads simulation state at test time.
The locked checkpoint is \texttt{litD4} (\texttt{ckpt\_pi\_lit\_distill\_v4}).

\subsection{Write--Use protocol}
\label{sec:write-use}

Each episode has two phases (Fig.~\ref{fig:protocol}).

\paragraph{Write (lit).}
A scripted geometric FSM runs for $T_w{=}25$ steps and stops at \texttt{descend\_A} (gripper above cube~A, before lift).
Stopping pre-lift leaves grasp$\rightarrow$lift$\rightarrow$place$\rightarrow$stack to Use instead of handing Use an in-air pose from the FSM.
We encode once: $z_0=\mathrm{Enc}(\mathrm{rgb}_w)$.

\paragraph{Use (dark).}
For $T_u$ steps (40 for the skill chain; 25 for HOLD), actions are $a_t=\pi(z_t)$.
Under \textbf{\darkf{}}, $z_{t+1}=f(z_t,a_t)$ with no new RGB.
Ablations: \texttt{lit\_reenc} re-encodes every frame; \texttt{freeze} holds $z_0$; \texttt{encode\_black} re-encodes black images.

\paragraph{Scoring.}
Rungs \texttt{near\_A}, \texttt{grasped}, \texttt{lifted}, \texttt{on\_B}, and \texttt{stacked} come from privileged state for metrics only; $\pi$ never sees them.
The headline metric is \texttt{stacked}; earlier rungs show where control fails.

\subsection{Ground-truth contact graph (eval only)}
\label{sec:gt-edges}

For contact probes we build edges from simulation state each step:
(gripper,A) if grasped or the end-effector is near A;
(A,B) if stacked or near in $xy$ while grasped;
(A,table) if A rests on the table.
\textbf{v2 masks} project entities through \texttt{base\_camera} onto image disks: neighbor masks zero GT-active regions; far masks zero small top corners outside those disks (Sec.~\ref{sec:gtcontact-v2}).
These masks are evaluation probes, not training labels.

\section{Experiments}
\subsection{Protocol and metrics}
\label{sec:protocol}

Unless noted, episodes follow Sec.~\ref{sec:write-use}: Write for $T_w{=}25$ to \texttt{descend\_A}, then Use with $a_t=\pi(z_t)$.
The skill chain uses $T_u{=}40$ under \darkf{} ($z_{t+1}=f(z_t,a_t)$, no re-encoding).
HOLD uses $T_u{=}25$ from the same pre-lift Write and scores approach, grasp, and lift during Use only.
The formal claim pools seed packs $\{0,1000\}$ with $n{=}80$ each ($n{=}160$ total).

\subsection{Main claim}
\label{sec:claim}

\begin{table}[t]
  \centering
  \caption{Skill-chain rung rates (\darkf{} vs.\ ablations, $n{=}160$, seed packs 0 and 1000).}
  \label{tab:claim}
  \footnotesize
  \begin{tabular}{lrrrrr}
    \toprule
    Mode & near\_A & grasped & lifted & on\_B & stacked \\
    \midrule
    lit\_reenc & 0.94 & 0.81 & 0.48 & 0.38 & 0.36 \\
    \darkf{} & \textbf{0.97} & \textbf{0.91} & \textbf{0.90} & \textbf{0.81} & \textbf{0.68} \\
    freeze & 0.93 & 0.02 & 0.00 & 0.00 & 0.00 \\
    \bottomrule
  \end{tabular}
\end{table}

Table~\ref{tab:claim} is the main result.
Early rungs (\texttt{near\_A}) look similar because Write already places the arm near cube~A.
The gap opens at \texttt{lifted}/\texttt{on\_B}/\texttt{stacked}: \darkf{} reaches \textbf{68.1\% stacked}, while \texttt{lit\_reenc} falls to 35.6\% and \texttt{freeze} collapses after grasp.
Re-encoding every step is therefore not free---without a consistent $f$ roll, Enc drift hurts late place geometry---and a frozen $z_0$ cannot finish the chain.
Optional ContactWrite variants on the same $n{=}160$ table yield stacked $63.8\%$ (\texttt{dark\_f\_cw}) and $68.1\%$ (\texttt{dark\_f\_cw\_gate}); we deploy plain \darkf{}.
\textbf{Seed-pack variance.} Pack~0 stacks at 73.8\% ($59/80$) and pack~1000 at 63.7\% ($51/80$).
A third pack (seed0${=}2000$, $n{=}80$, \darkf{} only) stacks at $58.8\%$, confirming pack-level variance without collapsing to chance.
The 68.1\% aggregate is not a single-seed fluke, but no pack clears a stable ${\geq}75\%$ bar.

\subsection{Write sensitivity}
\label{sec:write-sens}

\begin{table}[t]
  \centering
  \caption{Write sensitivity under \darkf{} ($n{=}80$ pooled over packs 0/1000; not the formal $n{=}160$ claim).}
  \label{tab:write-sens}
  \footnotesize
  \begin{tabular}{lrrrr}
    \toprule
    Setting & lifted & on\_B & stacked & $n$ \\
    \midrule
    $T_w{=}15$, stop \texttt{descend\_A} & 0.88 & 0.78 & 0.62 & 80 \\
    $T_w{=}25$, stop \texttt{descend\_A} & 0.88 & 0.78 & 0.62 & 80 \\
    $T_w{=}40$, stop \texttt{descend\_A} & 0.88 & 0.78 & 0.62 & 80 \\
    $T_w{=}40$, stop \texttt{reach\_above\_A} & 0.69 & 0.50 & 0.34 & 80 \\
    $T_w{=}40$, stop \texttt{close} & 0.99 & 0.84 & \textbf{0.68} & 80 \\
    \bottomrule
  \end{tabular}
\end{table}

Table~\ref{tab:write-sens} isolates the Write interface.
With stop fixed at \texttt{descend\_A}, stacked is \textbf{identical} for $T_w\in\{15,25,40\}$: success does not require a long lit budget once the FSM can reach the stop phase.
Changing the stop phase matters much more: stopping at \texttt{reach\_above\_A} drops stacked to $33.8\%$, while stopping later at \texttt{close} reaches $67.5\%$.
Write fragility is therefore about \emph{what} is encoded, not how many lit steps are allotted beyond that phase.
Write-time camera stress (Supp.): mild Gaussian noise barely moves stacked (${\approx}60$--$62\%$), whereas blur/JPEG at encode time cuts it to $17.5\%$ / $31.2\%$---consistent with Enc being the brittle handoff into dark Use.

\subsection{Failure modes}
\label{sec:failqual}

\begin{figure}[t]
  \centering
  \begin{subfigure}[t]{0.48\textwidth}
    \includegraphics[width=\linewidth]{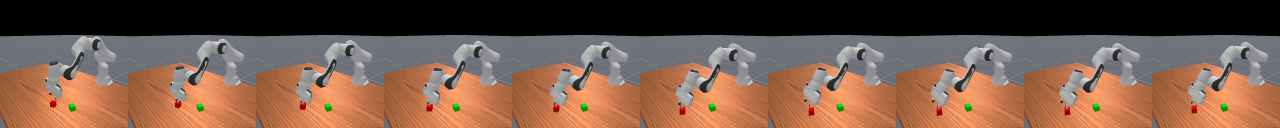}
    \caption{Pack~0: fail at \texttt{stacked} (seed~8). Late release/stabilize.}
  \end{subfigure}\hfill
  \begin{subfigure}[t]{0.48\textwidth}
    \includegraphics[width=\linewidth]{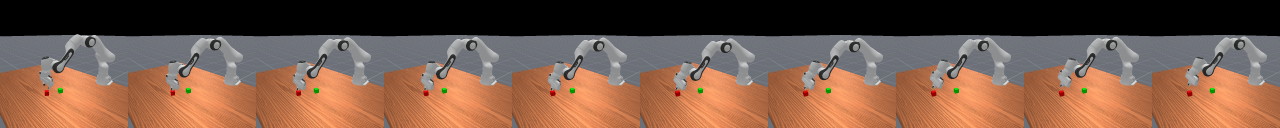}
    \caption{Pack~1000: fail at \texttt{on\_B} (seed~1001). Earlier place geometry.}
  \end{subfigure}
  \caption{\textbf{Failure qualitative} under locked \darkf{}.}
  \label{fig:failqual}
\end{figure}

We group \darkf{} failures by the first missed rung in each pack (Fig.~\ref{fig:failqual}).
On pack~0, most residual errors sit at \texttt{stacked} (13/80) and \texttt{on\_B} (5/80): grasp and lift often succeed, but release or stabilize fails.
On pack~1000, failures appear earlier (\texttt{near\_A}/\texttt{grasped}/\texttt{on\_B}), consistent with harder post-Write geometries under the same stop phase.
More gifs and strips are in the supplement (\texttt{fail\_qualitative/}).

\subsection{Mechanism}
\label{sec:mechanism}

\begin{figure}[t]
  \centering
  \begin{subfigure}[t]{0.48\textwidth}
    \includegraphics[width=\linewidth]{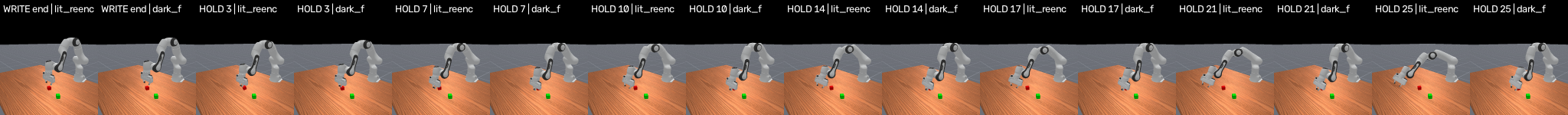}
    \caption{\textbf{Lit re-encode vs.\ \darkf{}} (seed~0). Left: per-step $\mathrm{Enc}(\mathrm{rgb})$; right: $z_0$ fixed with $f$ roll.}
  \end{subfigure}\hfill
  \begin{subfigure}[t]{0.48\textwidth}
    \includegraphics[width=\linewidth]{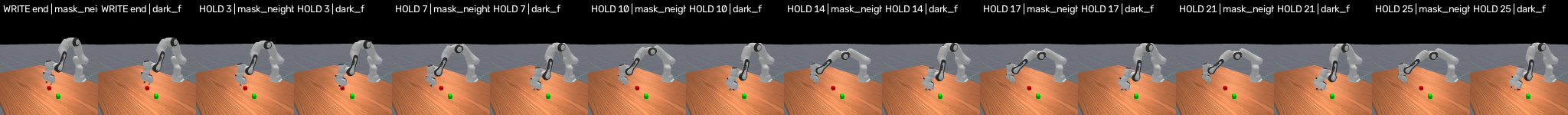}
    \caption{\textbf{GT neighbor mask vs.\ \darkf{}}. Left: projected contact-neighbor patches zeroed each re-encode; right: no re-encode.}
  \end{subfigure}
  \caption{\textbf{Mechanism rollouts} (contact sheets).}
  \label{fig:mechanism}
\end{figure}

\paragraph{Occlusion (HOLD).}
On HOLD ($n{=}40$), re-encoding under full or black occlusion yields ${\leq}43\%$ lift; \darkf{} reaches ${\approx}82.5\%$; freeze/scramble stay near $0\%$ (Fig.~\ref{fig:mechanism}, left).
Dark roll keeps pre-lift progress that lit re-encode loses when pixels are unreliable.

\paragraph{Dynamics.}
Offline, shuffle/true MSE on $f$ is $\mathbf{9.4\times}$; scrambling $z_p$ with true actions also raises error.
So $f$ is neither an identity nor action-agnostic.

\paragraph{GT contact influence.}
Edges \texttt{gripper--A}, \texttt{A--B}, and \texttt{A--table} come from simulation state.
Coarse v1 masks zero lit lift for both neighbor and irrelevant proxies, while \darkf{} gains $+0.40$ over full re-encode.
Camera-projected v2 patches (Fig.~\ref{fig:gt-overlay}) appear in Table~\ref{tab:gtcontact} and Sec.~\ref{sec:gtcontact-v2}.

\subsection{Fair nulls}
\label{sec:nulls}

All nulls share the lit Write and differ only in Use.
Kernel \darkf{} reaches 68\% stacked ($n{=}160$); untrained latent-action stubs stay near $0\%$ on the same rungs.
State-RSSM one-step MPC reads \emph{privileged} geometry in DARK (${\approx}98\%$ HOLD lift, ${\approx}35\%$ stacked).
We treat it as a stronger-observation upper bound, not a matched visual baseline: HOLD lift should not be compared when DARK inputs differ.

\subsection{Vision Write and matched pixel baselines}
\label{sec:fair-write}

Privileged FSM Write is the locked claim interface.
We additionally report \textbf{vision Write}: RGB-only actions from a dedicated $\pi_{\mathrm{write}}$ (Enc/$f$/$\pi_{\mathrm{use}}$ frozen; BC on pre-lift expert).
On matched packs ($n{=}80$), vision-budget$+$\darkf{} reaches \textbf{35.0\%} stacked (vs.\ 26.3\% when reusing $\pi_{\mathrm{use}}$ for Write; FSM$+$\darkf{} 62.5\%; freeze ${\approx}0$).
A brightness/blackout detector enables deploy-style dark handoff (auto-switch 32.5\% stacked vs.\ always-lit ${\approx}0$ after injected blackout; $n{=}40$).

\begin{table}[t]
  \centering
  \caption{Fair Use after matched Write ($n{=}80$, packs 0/1000; $T_w{=}25$, $T_u{=}40$).
  Dreamer-style / pixel RSSM use no 9D GT; state-RSSM does (unfair visual match).}
  \label{tab:fair}
  \footnotesize
  \setlength{\tabcolsep}{3pt}
  \begin{tabular}{llrr}
    \toprule
    Write & Use & stacked & note \\
    \midrule
    FSM & \darkf{} (kernel) & 0.625 & claim interface \\
    vision+$\pi_{\mathrm{write}}$ & \darkf{} & \textbf{0.350} & anti-cheat \\
    vision+$\pi_{\mathrm{write}}$ & freeze & 0.000 & \\
    FSM & Dreamer-style dark & 0.000 & fair; not DreamerV3 \\
    FSM & pixel-RSSM prior & 0.000 & fair visual null \\
    FSM & state-RSSM MPC & ${\approx}0.35$ & reads 9D in dark \\
    \bottomrule
  \end{tabular}
\end{table}

A Dreamer-\emph{style} PyTorch agent (RSSM$+$BC$+$imagination actor--critic; \emph{not} official DreamerV3) and a lite pixel RSSM both score \textbf{0\%} stacked under the same card (including lit re-encode).
We report them as observation-matched visual nulls, not as DreamerV3/DreamZero SOTA comparisons (those need a dedicated Write--Use adapter).

\subsection{Appearance OOD}
\label{sec:appearance}

Write-time retexture of $\mathrm{Enc}(\mathrm{rgb})$ leaves warm/cool stable but collapses dusk/night (night stacked ${\rightarrow}0$).
With write fixed to \texttt{default} and OOD only in HOLD, \darkf{} lift stays ${\approx}83\%$ for dusk/night while lit\_reenc drops to ${\approx}0\%$, separating write-time encoding fragility from hold-time roll robustness.
Enc-only appearance DR (dusk/night in training, $\pi/f$ frozen) neither restores night write nor preserves default stacked (${\sim}5\%$ vs.\ 68\%); $\pi$ and $f$ would need to be co-trained (Limitations).

\subsection{Second domain}
A PushCube privileged oracle reaches 95\% success ($n{=}40$).
We do not cross-evaluate StackCube \texttt{litD4} on Push; the row only shows that a second ManiSkill domain is instrumented.

\subsection{GT contact v2 (projected patches)}
\label{sec:gtcontact-v2}

\begin{figure}[t]
  \centering
  \includegraphics[width=0.95\linewidth]{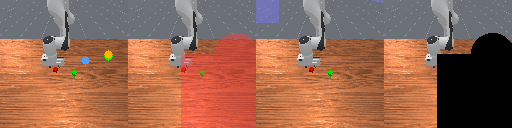}
  \caption{\textbf{GT contact v2 masks} (seed~0, write end): entity projections; \textbf{neighbor} (red) vs \textbf{far corners} (blue); masked RGB. Mask precision/recall vs GT neighbor region = 1.0 (Table~\ref{tab:gtcontact}).}
  \label{fig:gt-overlay}
\end{figure}

We test whether GT contact neighborhoods are needed for lit re-encoding on HOLD, and whether \darkf{} already carries contact-relevant progress without those pixels (Fig.~\ref{fig:gt-overlay}).
Each HOLD step uses the contact graph in Sec.~\ref{sec:gt-edges}.
\textbf{Neighbor masks} zero camera-projected disks on entities in active GT edges; \textbf{far masks} zero small top corners outside those disks.
Against the GT neighbor region, mask precision and recall are both 1.0 (v2), unlike coarse v1 bands.

\begin{table}[t]
  \centering
  \caption{GT contact v2 on HOLD ($n{=}40$, patch radius 22px).}
  \label{tab:gtcontact}
  \footnotesize
  \begin{tabular}{lrrrr}
    \toprule
    Condition & lift & $\Delta$lift & mask P & mask R \\
    \midrule
    full & 0.425 & 0.000 & -- & -- \\
    mask\_gt\_neighbor & 0.000 & $-$0.425 & 1.00 & 1.00 \\
    mask\_gt\_far & 0.075 & $-$0.350 & -- & -- \\
    \darkf{} & 0.825 & $+$0.400 & -- & -- \\
    \bottomrule
  \end{tabular}
\end{table}

\paragraph{Reading.}
Neighbor masks collapse lit lift ($0.43{\rightarrow}0$), while far masks are milder ($0.08$), so $|\Delta\mathrm{lift}|_{\mathrm{neighbor}} > |\Delta\mathrm{lift}|_{\mathrm{far}}$.
\darkf{} \emph{raises} lift by $+0.40$ over full re-encode: contact-relevant control lives in write-time $z$ and open-loop $f$, not in per-step contact pixels.
The HOLD neighbor--far $|\Delta$lift$|$ gap is only $0.075$; the main signal is neighbor collapse versus \darkf{} survival.

\paragraph{Skill chain.}
Supplementary Table~S1 reports full-chain GT masking ($n{=}40$): neighbor and far both drive stacked from $0.40$ to $0$, while \darkf{} stays at 67.5\% ($+0.28$ vs.\ full re-encode).

\section{Related Work}
\label{sec:related}

\paragraph{World models and JEPA-style representations.}
Predictive and joint-embedding methods study representation quality under occlusion~\citep{lecun2022path,assran2023self,bardes2024revisiting}.
PHYRE popularized qualitative attention probes for physical reasoning~\citep{bakhtin2019phyre}; video JEPA work often emphasizes future prediction.
We instead fix a Write--Use control protocol, score skill rungs, and measure contact influence with camera-projected masks and precision/recall.

\paragraph{Dreamer / RSSM.}
Dreamer-style agents learn latent dynamics for imagination and control~\citep{hafner2019learning,hafner2020dreamer,hafner2023mastering}.
Our state-RSSM null reuses the same lit Write but reads privileged 9D geometry with one-step MPC in Use (high HOLD lift, low stacked).
Observation-matched Dreamer-\emph{style} / pixel RSSM agents (no 9D GT) score 0\% stacked under the same card; we do not claim an official DreamerV3 or DreamZero comparison without a dedicated adapter.
The privileged RSSM row highlights DARK information asymmetry rather than matched visual dominance.

\paragraph{Imitation and visuomotor policies.}
Behavioral cloning and implicit policies~\citep{florence2020implicit}, sequence models such as ACT~\citep{zhao2023act}, and diffusion policies~\citep{chi2023diffusionpolicy} improve imitation when RGB is available every step.
Time-contrastive visual imitation likewise targets robust features under continuous observation~\citep{sermanet2018time}.
Our setting is complementary: dark Use after a single encode.
Untrained latent-action stubs under identical rungs show that the protocol alone does not produce stacked success.

\paragraph{Simulation benchmarks.}
StackCube experiments use ManiSkill~\citep{gu2023maniskill2,ta2023maniskill3}.
A PushCube oracle row documents a second domain; it is not zero-shot transfer of the StackCube checkpoint.

\section{Discussion}
\label{sec:discussion}

Two practical points follow from the locked protocol.
First, \textbf{dark Use is viable on StackCube} when Write yields a usable $z_0$: \darkf{} beats continuous re-encoding on late rungs, and occlusion/GT-mask probes show that contact-relevant progress need not be refreshed from pixels every step.
Second, \textbf{Write is the brittle interface}: stacked is insensitive to $T_w$ once the stop phase is reached, but earlier stops, write-time blur/JPEG, and dusk/night retexture all wipe out late rungs, while the same appearance shifts during HOLD do not hurt \darkf{} lift.
Improving Enc under OOD therefore requires co-adapting $\pi$ and $f$; Enc-only DR was not enough in our trial.
Fair comparison also requires matched Write and matched DARK information: privileged RSSM HOLD lift is not evidence against visual dark roll when stacked rates already favor the kernel under weaker Use observations.

\paragraph{Few-shot transfer and habit vs.\ physics.}
A companion paper~\citep{hang2026habit} asks why freezing shared dynamics and adapting only a thin command map $g$ is preferable when demonstration multimodality largely reflects action/habit choice rather than multiple physical laws.
\textbf{Primary attribution} is in StackCube physical latent $z_p$ (one-step mix/true MSE ${\sim}10.4\times$; same-$z_0$ rollouts); toy explicit state and DROID proprio provide sanity checks and robot-side diagnosis (wrong-action spread ${\sim}13\times$ vs.\ true targets; idle segments narrow).
On the \emph{same} pretrained $f^*$, N-shot \textbf{freeze $f$ + thin $g$} beats scratch and is competitive with full finetune at low $N$ on both kernel and proprio heads; multistep $H{=}5$ rollouts agree.
That line does \textbf{not} re-derive latent action $\tilde a$ (cf.\ AdaWorld).
DreamZero LoRA in the companion appendix is an \textbf{external deployment baseline}, not the proposed method---it does not isolate the Physical Kernel factorization and, under matched $N$, did not improve open-loop action MSE over the pretrained checkpoint at 150 steps.
Full evidence ladder, figures, and reproduction are in the standalone companion~\citep{hang2026habit}; this manuscript remains focused on Write--Use skill chains under \darkf{}.

\section{Limitations}
\label{sec:limitations}

\begin{itemize}
  \item \textbf{Aggregate success.} Stacked rate is 68.1\% overall (73.8\% / 63.7\% on packs 0 / 1000; 58.8\% on pack 2000, $n{=}80$); we do not claim a stable ${\geq}75\%$ gate.
  \item \textbf{Privileged Write.} The FSM remains the strongest Write; a dedicated $\pi_{\mathrm{write}}$ reaches 35\% vision-budget stacked ($n{=}80$) but not FSM parity---the primary claim still concerns Use under \darkf{}.
  \item \textbf{GT contact v2.} Edges and disks are heuristic; the HOLD neighbor--far $|\Delta$lift$|$ gap is modest ($0.075$). Skill-chain GT masking appears in Supp.\ Table~S1 as a small-$n$ probe.
  \item \textbf{Baselines.} Privileged RSSM is not observation-matched; Dreamer-style/pixel nulls are 0\% under the fair card and are \emph{not} official DreamerV3/DreamZero. This draft has no hardware results yet (pilot planned).
  \item \textbf{Write-time appearance.} Night write still collapses stacked on \texttt{litD4}. Enc-only DR (25 epochs; dusk/night in train; $\pi/f$ frozen) leaves night ${\sim}5\%$ and drops default stacked from 68\% to ${\sim}5\%$ ($n{=}40$), indicating encoder drift without co-training $\pi$ and $f$.
\end{itemize}

\section{Conclusion}
\label{sec:conclusion}

We studied dark manipulation on ManiSkill StackCube: after a short lit Write, \darkf{} runs $\pi(z)$ and $f(z,a)$ through a full skill chain without new RGB.
The locked recipe reaches 68.1\% stacked ($n{=}160$), with seed-pack variance and evidence that $f$ is action-sensitive and that dark roll survives occlusion and GT contact masking where lit re-encode fails.
Write sensitivity shows that stop phase---not lit length---gates usable $z_0$; appearance OOD and encode-time blur/JPEG likewise localize fragility at write-time encoding, not hold-time roll.
Fair nulls clarify that privileged DARK observers are not matched visual baselines; vision Write with $\pi_{\mathrm{write}}$ and observation-matched Dreamer-style/pixel nulls further bound the claim.
Future work includes co-trained appearance DR, stronger vision Write, official DreamerV3/DreamZero adapters under the same card, and real-robot validation.

\bibliographystyle{plain}
\bibliography{refs}

\end{document}